\documentclass[10pt,twocolumn,letterpaper]{article}

\usepackage{cvpr}
\usepackage{times}
\usepackage{epsfig}
\usepackage{graphicx}
\usepackage{amsmath}
\usepackage{amssymb}
\usepackage{float}
\usepackage{multirow}
\usepackage{multicol}
\usepackage{array}
\usepackage{subfigure}
\usepackage[dvipsnames]{xcolor}
\colorlet{Mycolor1}{green!10!orange!90!}

\def\etal{\emph{et al.~}}
\def\ie{\emph{i.e.}}
\newcommand{\tabincell}[2]{\begin{tabular}{@{}#1@{}}#2\end{tabular}}  
\usepackage[T1]{fontenc}
\usepackage[pagebackref=true,breaklinks=true,colorlinks,bookmarks=false]{hyperref}

\cvprfinalcopy 

\ifcvprfinal\fi
\begin{document}

\title{Towards Robust RGB-D Human Mesh Recovery}

\author{Ren Li$^{1}$, Changjiang Cai$^{2}$, Georgios Georgakis$^{3}$, Srikrishna Karanam$^{1}$, Terrence Chen$^{1}$, and Ziyan Wu$^{1}$\\
$^{1}$United Imaging Intelligence, Cambridge MA\\
$^{2}$Department of Computer Science, Stevens Institute of Technology, Hoboken NJ \\
$^{3}$Department of Computer Science, George Mason University, Fairfax VA\\
{\tt\small \{first.last\}@united-imaging.com,ccai1@stevens.edu,ggeorgak@gmu.edu}
}

\makeatletter
\g@addto@macro\@maketitle{
\vspace{-15 pt}
\begin{figure}[H]
  \setlength{\linewidth}{\textwidth}
  \setlength{\hsize}{\textwidth}
  \centering
  \includegraphics[width=17cm]{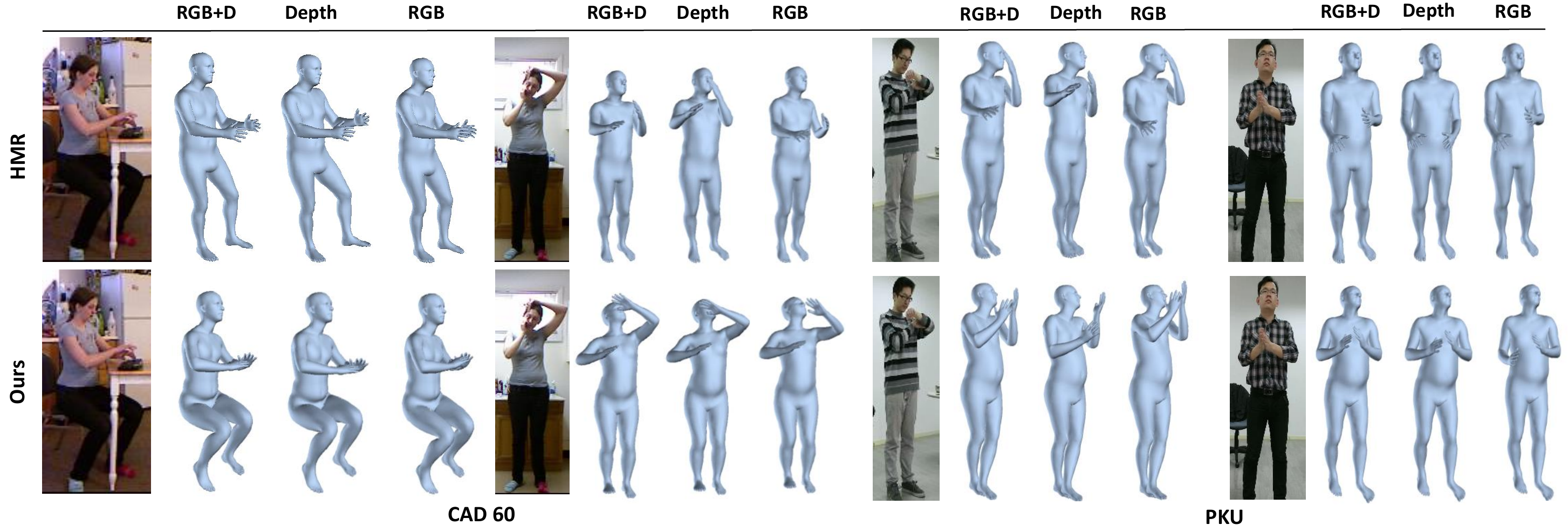}
  \caption{Results on the CAD 60 and PKU-MMD dataset. Our framework includes a new dynamic data fusion module that enables our model to recover human mesh from any of RGB-only, depth-only or RGB-D inputs. We also propose a universal SMPL constraint generator that provides for SMPL constraints when explicit annotations are not available. This facilitates principled model training in conjunction with our novel depth ranking consistency learning objective. First row: baseline HMR \cite{kanazawaHMR18}, second row: our results.}\label{fig:results}
\end{figure}
}
\makeatother
  
\maketitle
\thispagestyle{empty}

\begin{abstract}
We consider the problem of human pose estimation. While much recent work has focused on the RGB domain, these techniques are inherently under-constrained since there can be many 3D configurations that explain the same 2D projection. To this end, we propose a new method that uses RGB-D data to estimate a parametric human mesh model. Our key innovations include (a) the design of a new dynamic data fusion module that facilitates learning with a combination of RGB-only and RGB-D datasets, (b) a new constraint generator module that provides SMPL supervisory signals when explicit SMPL annotations are not available, and (c) the design of a new depth ranking learning objective, all of which enable principled model training with RGB-D data. We conduct extensive experiments on a variety of RGB-D datasets to demonstrate efficacy. 
\vspace{-20 pt}
\end{abstract}

\section{Introduction}\label{sec:intro}
We consider the problem of estimating a parametric human mesh model for human pose estimation. Given a test image of a person and a known parametric mesh model, \eg, the 3D SMPL model \cite{SMPL:2015}, we seek to estimate the parameters of the model that best explain the human pose in the test image. This problem finds applications in a diverse set of areas including healthcare \cite{obdrvzalek2012accuracy} and robotics \cite{martinez2018real,zimmermann20183d}, among others \cite{belghit2018vision,kadkhodamohammadi2017multi,kohli2013key,moeslund2006survey}. There has been substantial progress in the recent past in this area, with  \cite{alp2018densepose,cao2017realtime,kanazawaHMR18,wei2016cpm,Zhou_2017_ICCV} providing a representative overview. 
Most existing deep human pose estimation algorithms solely attempt to recover the mesh parameters from a single RGB image, which is an inherently under-constrained problem since there are many possible 3D configurations that can in principle explain the same 2D projection. Consequently, these methods, typically trained only with 2D keypoints or SMPL annotations, are unable to resolve these ambiguities by design. While SMPL annotations do provide stronger supervisory signals when compared to using only 2D keypoints, one is still attempting to regress these 3D SMPL parameters directly from an RGB image, with no notion of the input depth that is key to resolving this depth ambiguity problem. As we demonstrate in this work, using depth data that is aligned with RGB images, which is easy to obtain with the advent of affordable consumer-focused sensors such as the Kinect line of devices \cite{kohli2013key}, helps address this problem, resulting in reduced mesh reconstruction errors. 

Using aligned RGB-D data to train a human mesh predictor with existing approaches would need the availability of datasets that are annotated with both keypoints as well as SMPL parameters. However, as noted above, since much recent progress has been in the RGB-only domain, there are no publicly available RGB-D datasets that provide both keypoints and SMPL annotations for training, with the few available ones providing only keypoints information. Next, the RGB-D datasets that do have keypoints annotations may vary in definitions (\eg, the coordinate systems in which the keypoints are defined). Finally, the few RGB-only datasets that have these annotations do not come with aligned depth data \cite{VNect_SIGGRAPH2017}. Consequently, we ask two key questions in this work- (a) how can we train mesh estimators with such datasets that provide different kinds of annotations? and (b) how can we generate universal 3D supervisory signals from RGB-D datasets that do not have SMPL annotations available?

In this work, we take a step towards solving the key depth ambiguity problem by addressing the aforementioned questions in a principled manner. To tackle the challenge posed by the lack of aligned depth data (or even RGB data) in some datasets, we propose a new dynamic data fusion module, along with an associated training policy, that facilitates the functioning of the trained model with only RGB inputs, only depth inputs, or both. To realize this functionality, our training procedure includes a randomly activated data stream selection functionality that selects one of the two input stream (RGB or depth) feature representations when optimizing the overall learning objective. This results in a model trained to work with any of the three resulting combinations of data availability. Next, to tackle the challenge posed by RGB-D datasets having 3D keypoints information in different coordinate systems, and the lack of SMPL annotations, we propose a new module we call the \textsl{Universal SMPL Constraint Generator} (USCG). Our USCG is designed to be agnostic to the specific coordinate system in which the 3D keypoints may be defined. This is achieved by explicitly enforcing USCG to only look at the relative, root-agnostic, configuration of the 3D keypoints as input. More crucially, the USCG is trained to output auxiliary SMPL parametric information that we use to generate 3D SMPL supervisory signals for training our model in the absence of explicit SMPL annotations. Finally, our training procedure involves a new learning objective we call the \textsl{depth ranking consistency} (DRC) loss. The proposed DRC loss ensures the mesh parameters estimated by our model respects the relative configuration of the predicted keypoints on the depth data (\eg, one is closer than the other). This is achieved by means of explicit depth ranking supervision on the ordering of the predicted keypoints. As indicated in Figure \ref{fig:results}, these components in together help us to generate more reliable and accurate results of human mesh recovery.

To summarize, the key contributions of this work include:

\begin{itemize}
    \item We address the problem of depth ambiguity in existing RGB-based human mesh recovery algorithms by proposing three new functionalities. 
    \item We propose a new dynamic data fusion module to facilitate training human mesh predictors with datasets that do not have aligned RGB-D data.
    \item We propose a new constraint generator that provides auxiliary SMPL supervisory signals for RGB-D datasets that do not have SMPL annotations, and is agnostic to the coordinate system in which their keypoint annotations may be defined. 
    \item We propose a new depth ranking consistency learning objective that ensures the model estimates mesh parameters that result in respecting the relative depth ordering of predicted keypoints. 
\end{itemize}

\begin{figure*}[!htbp]
  \centering
\includegraphics[width=17cm]{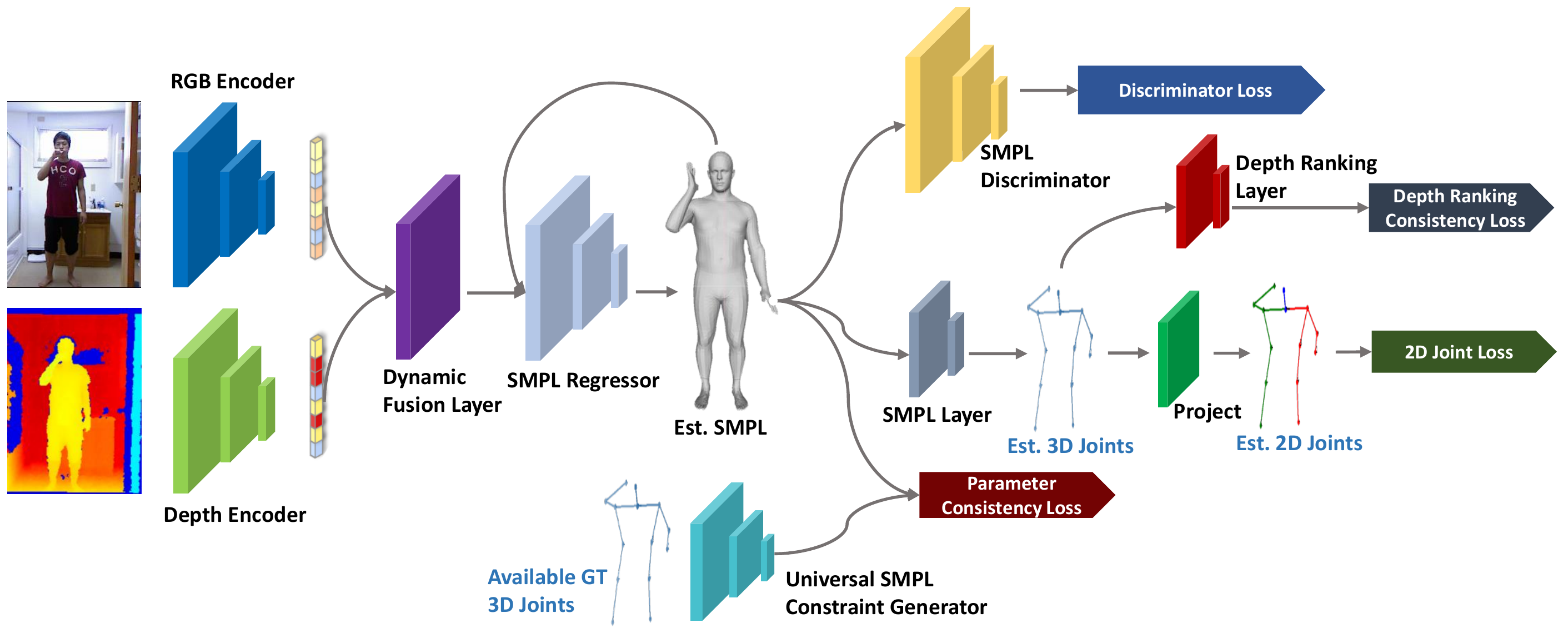}
\caption{The pipeline of the proposed depth-aware human pose regression model. Our model takes RGB images and depth maps as input. After the dynamic fusion of RGB-D features, the SMPL regressor regresses the human body parameters. We use different kinds of consistency to constrain the regression. }\label{fig:pipeline}
\end{figure*}

\section{Related Work}\label{sec:related}
Human pose estimation has been extensively studied in the past and we refer readers to \cite{moeslund2001survey,moeslund2006survey,sarafianos20163d} for detailed surveys. Here, we review algorithms that are most immediately related to our work, \ie, estimating parametric human mesh models and 3D human pose estimation, helping differentiate our method while also putting it in proper context. 

\noindent \textbf{RGB-only 3D pose estimation from 2D.} \, Much effort has been expended in estimating the 3D human pose given 2D keypoints information. Martinez \etal \cite{martinez_2017_3dbaseline} proposed several simple techniques, \eg, estimating 3D joints in camera frame and baseline architectural changes such as batch normalization, that resulted in performance improvements of existing baseline algorithms that ``lift" 2D joints to 3D. Wandt and Rosenhahn \cite{Wandt2019RepNet} proposed to perform this lifting by means of a distribution mapping approach that is trained to map a given 2D pose to its most likely 3D counterpart. Jahangiri \etal \cite{jahangiri2017generating} proposed to use a multi-modal mixture density network to generate multiple plausible 3D pose hypotheses, all of which can explain the input 2D joints. To address this 2D-3D ambiguity problem, several constrained search methods have been proposed. Jiang \cite{jiang20103d} used a nearest neighbor search strategy to attempt to resolve the ambiguity, whereas Gupta \etal \cite{gupta20143d} and Chen and Ramanan \cite{chen20173d} explored variants of a temporal search strategy. Some more recent end-to-end 3D pose estimators require large amounts of annotated, and in possibly paired, 2D and 3D data. Pavlakos \cite{pavlakos2017volumetric} extended the stacked Hourglass \cite{newell2016stacked} model to a coarse-to-fine volumetric prediction in 3D space. Sun \etal \cite{sun2018integral} used the differentiable \textit{soft-argmax} \cite{bahdanau2014neural,levine2016end} operation to unify heatmap representation and joint regression, giving a flexible approach compatible with existing heatmap-based algorithms. Given that approches along this direction require (costly) large-scale annotations, there have also been efforts in learning from synthetic data \cite{chen2016synthesizing,deprelle2019learning,groueix2018b,peng2018jointly,rogez2016mocap,varol17_surreal}, or  training algorithms in un-/self-/weakly-supervised ways \cite{chen2019weakly,kocabas2019epipolar,rhodin2019neural,rhodin2018unsupervised, Zhou_2017_ICCV}. All these techniques rely on using only RGB images in estimating the 3D pose, whereas we use the aligned depth image as well in conjunction with a model training strategy that explicitly attempts to address the depth ambiguity problem. 

\noindent \textbf{RGB-only 3D Human Mesh Recovery.} \, With the simplicity and extensibility of the SMPL model \cite{SMPL:2015}, there has been substantial recent work in estimating the parameters of this model. Bogo \etal \cite{Bogo:ECCV:2016} fit the model by minimizing the reprojection error between the 3D model joints and the input 2D joints. Kannzawa \etal \cite{kanazawaHMR18} presented the first end-to-end version of this strategy with an adversarial training scheme, taking as input a single RGB image and outputing the model parameters in one inference step. Some recent extensions of this work included a graph-CNN strategy \cite{kolotouros2019convolutional} for directly regressing the vertices of the SMPL model, exploiting video temporal context \cite{arnab2019exploiting} for reducing the 2D-3D ambiguity, and directly learning human motion dynamics \cite{kanazawa2019learning} from input video. In contrast, our approach relies on RGB-D data to estimate the parameters of the SMPL model, where a key difference to existing work is we do not rely on the availability of explicit SMPL annotations for training.

\noindent \textbf{Depth-based 3D Pose Estimation.} \ 
While not as popular as RGB-only approaches, mostly due to the lack of relavant annotations (\eg, SMPL data), there is some prior work in using depth data for 3D pose estimation. Kadkhodamohammadi \etal \cite{kadkhodamohammadi2017multi} proposed a multi-view RGB-D approach for human pose estimation in surgical operating rooms, demonstrating the benefit of using depth data in reducing pose ambiguity. Pavlakos \etal \cite{pavlakos2018ordinal} proposed using ordinal depth constraints as a form of weak supervision, training a model to predict the 3D pose. Such constraints were also used by Sharma \etal \cite{Sharma_2019_ICCV} to score and refine candidate 3D poses. In contrast, we propose a dynamic fusion module that enables training models with RGB and RGB-D data having a wide variety of annotations to address the ambiguity problem, while also explicitly recovering the human mesh which was not part of the design of these methods.

\section{Approach}
As noted in Section \ref{sec:intro}, existing approaches that perform parametric human mesh estimation rely only on RGB images, thereby solving an inherently under-constrained problem by design. To address this problem, we propose to explicitly use depth data aligned with the corresponding RGB images, potentially providing an additional source of supervisory information. However, a naive extension of existing approaches, like \eg, Kanazawa \etal \cite{kanazawaHMR18}, is sub-optimal since accurately training such a model will need the availability of keypoints and SMPL annotations. While some human RGB-D datasets do have keypoints information, they most certainly do not have SMPL annotations. On the other hand, the few RGB-only datasets that do have SMPL annotations in addition to keypoints information do not come with aligned depth data. To address this crucial problem, our proposed pipeline (see Figure~\ref{fig:pipeline}) comprises several important components- (a) a novel dynamic data fusion module that enables the trained model to work as well with RGB-only, depth-only, or RGB-D data as input, (b) a novel universal SMPL constraint generator (USCG) module that helps generate auxilialy 3D SMPL supervisory signals for RGB-D datasets that do not have explicit SMPL annotations, and (c) various learning objectives, of which a new depth ranking loss ensures the trained model respects the relative configuration or ordering of the predicted keypoints on the depth map. In this section, we describe our key innovations in detail. We first start with a brief review, in Section~\ref{sec:smplReview} of the SMPL model, the parametric human mesh model we use in this work. We then describe our proposed dynamic data fusion method in Section~\ref{sec:dynamicFusion}, the USCG in Section~\ref{sec:uscg}, and the ranking loss in Section~\ref{sec:rankLoss}.

\subsection{3D Human Body Representation}
\label{sec:smplReview}
We use the Skinned Multi-Person Linear (SMPL) model of Loper \etal \cite{SMPL:2015} to represent the human body. The SMPL model is a statistical parametric differentiable model that operates on shape parameters $\beta \in \mathbb{R}^{10}$ and pose parameters $\theta \in \mathbb{R}^{72}$. Given the shape $\beta$ and the pose $\theta$, the SMPL model defines a function $M(\beta,\theta,\Phi): \mathbb{R}^{82}\rightarrow \mathbb{R}^{3N}$ that produces the N mesh vertices representing the human body. The shape parameters $\beta$ comprise the first 10 coefficients of a principal component analysis (PCA) space, responsible for outputting a blend shape sculpting the identity. Pose in the SMPL model is represented using the standard skeletal rig and is parameterized by the joint angles, in axis-angle representation, of each of the 23 joints of the rig as well as the root joint. This gives $3\times23 + 3 = 72$ parameters to represent the pose in the SMPL model, accounting for mesh deformations that are dependent on the pose of the joints. Given a particular $\beta$ and $\theta$, and the learned model parameters $\Phi$, one can compute the 3D joints locations $X(\beta,\theta; \Phi)$ from a pre-defined linear regression of the mesh vertices, and we define this particular operation as the SMPL layer in our pipeline. 

Given the 3D joints locations, we follow Kanazawa \etal \cite{kanazawaHMR18} to project them to the 2D image plane and obtain 2D image keypoints. Specifically, we use the weak-perspective camera model defined as:
\begin{equation}\label{eq:projection}
    x=s\Pi(RX(\beta,\theta)) + t,
\end{equation}
where $R\in \mathbb{R}^3$ is the global rotation in axis-angle representation, $t\in \mathbb{R}^2$ and $s\in \mathbb{R}$ are translation and scale, and $\Pi$ is an orthographic projection. Therefore, the pose of the human body in the image is represented by $\Theta=\{\theta,\beta,R,s,t\}$. 


\subsection{Dynamic Data Fusion}
\label{sec:dynamicFusion}
A simple, straightforward way to use both RGB and depth images for mesh recovery is to concatenate the RGB and depth features before inferring the mesh parameters, which can be done using any existing RGB-only based parameter inference pipeline. In fact, this is indeed one of the baselines we experimentally compare our proposed approach with later in the paper. For this baseline, we use the recent work of Kanazawa \etal \cite{kanazawaHMR18}, where the feature encoder is followed by an iterative SMPL regressor that outputs the pose and shape parameters.  

However, as noted previously, not all RGB-D datasets come with both 2D and 3D annotations for us to use the simple fusion scheme described above. In fact, many of the human RGB-D datasets that are relevant in our context \cite{pku-mmd,cad60} only have 3D skeletal joints annotations, whereas several RGB-only pose estimation datasets have mostly 2D keypoints annotations \cite{lsp,coco}. Consequently, one naturally asks the question- given these datasets with various kinds of annotations, how can we train a parametric mesh estimator that makes the best use of all the available data? As an answer to this challenge, we propose a novel dynamic data fusion module, which we show can facilitate dynamical fusion of features, before parameter estimation, irrespective of whether the input is RGB-only, depth-only, or RGB-D.

As shown in Figure \ref{fig:pipeline}, we first process the RGB and depth data using two separate convolutional encoder (we use ResNet-50 \cite{he2016deep,resnet}) to extract the RGB features $\phi_{RGB}$ and depth features $\phi_{D}$. We perform all fusion operations in the feature space, and to this end, we employ a feature concatenation unit to concatenate the feature vectors $\phi_{RGB} \in \mathbb{R}^{d}$ and $\phi_{D} \in \mathbb{R}^{d}$, and process the concatenated feature representation $[\phi_{RGB}, \phi_{D}]$ with two additional fully connected units to obtain the final fused feature representation $\phi_{fuse}$. This is the baseline fusion strategy. 

However, given that aligned RGB and depth data are not always readily available, we simulate this aspect using a new probabilistic training policy that attempts to address this missing data problem. Specifically, during training, we add random noise to features extracted from the RGB and the depth streams. To realize this, we randomly select one of the RGB and depth streams with a probability $p_{\text{miss}}$ ($p_{\text{miss}}<0.5$), which means there is a case that neither of the streams is chosen. We then replace the original input of the selected stream with an input array comprising all zeros as pixel values, and concatenate the feature representation of this stream $\phi_{void}$ and the feature representation of the other stream. In this way, during the course of training our network, the fusion module will encounter three kinds of feature representations as input: $[\phi_{RGB},\phi_{D}]$, $[\phi_{RGB},\phi_{void}]$, and $[\phi_{void},\phi_{D}]$, simulating the three situations where both RGB and depth data are present, and one of the two is absent. Our network will then be trained as usual with this dynamically fused feature representation. During inference, if both RGB and depth data are present, we use both as input to our network; otherwise the missing stream will be substituted with blank images as above. However, since our model was trained using the policy described above, it will have learned to fuse features dynamically according to the presence of $\phi_{void}$.

\subsection{Universal SMPL Constraint Generator}
\label{sec:uscg}
As shown in Kanazawa \etal \cite{kanazawaHMR18} and Liang \etal \cite{liang2019shape}, having data with SMPL annotations is crucial for learning an accurate human mesh estimator. However, most RGB-D datasets \cite{pku-mmd,cad60} that are relevant in our context do not have SMPL annotations, and only make available the 3D keypoints locations. Moreover, each RGB-D dataset will have its own definition of the coordinate system in which these keypoints are defined, further exacerbating training a mesh estimator on multiple datasets jointly. As an answer to this challenge, we propose a new module we call the universal SMPL constraint generator (USCG), whose goal is to unify the keypoints annotations across multiple datasets and generate auxiliary SMPL constraints which we then use to further regularize our mesh estimator learning process. Simply put, our proposed USCG modules removes the need for obtaining SMPL annotations for these RGB-D datasets, which in itself is a highly complex process. In fact, as noted by Kanazawa \cite{kanazawaHMR18}, this is an elaborate process involving using ground-truth mocap data (relevant RGB-D datasets noted above do not have this data).

\begin{figure}[htbp]
  \centering
\includegraphics[width=8cm]{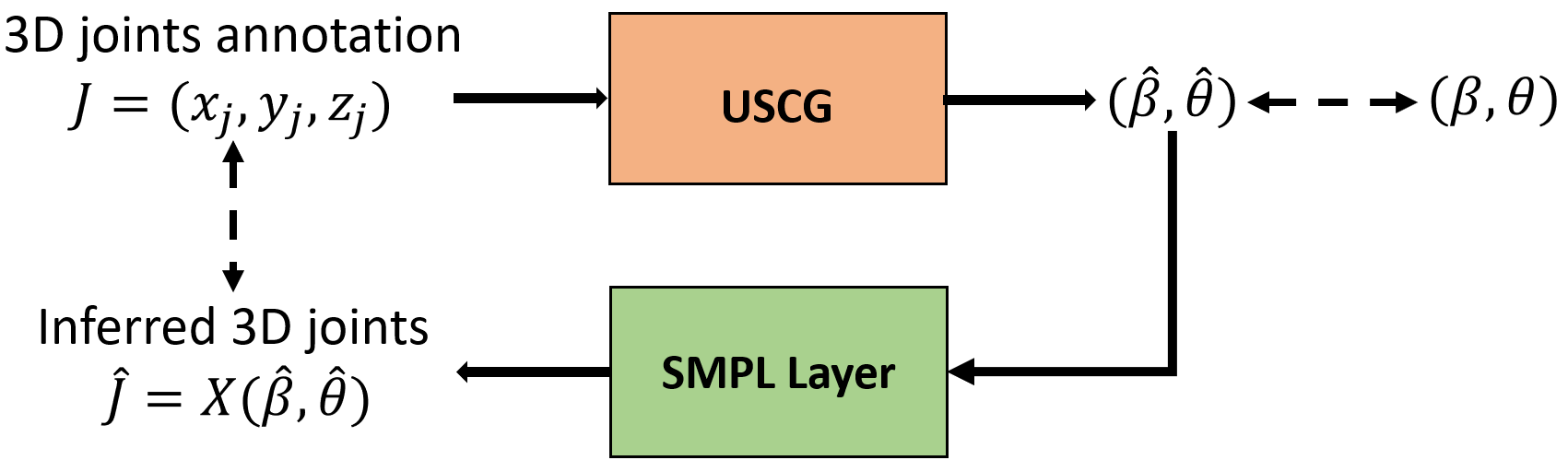}
\caption{Our proposed USCG.}\label{fig:USCG}
\end{figure}

To this end, we train our USCG module to take as input a set of 3D keypoints locations and produce the corresponding SMPL parameters $\beta$ and $\theta$. One key aspect of our USCG is that the input set of keypoints is always relative to the root location, \ie, the root location is set to origin in a reference frame common to all the RGB-D datasets. This way, once trained, the USCG can be used to process 3D keypoints from any given RGB-D dataset. To train the USCG we need paired 3D keypoints and SMPL parameter data. A large number of such paired data can be easily obtained from the synthetically created data from Loper \etal \cite{loper2014mosh} from the raw 3D MoCap markers. The USCG training process is visually summarized in Figure~\ref{fig:USCG}. Given 3D keypoints information $J$, the USCG $G$ estimates $[\hat{\beta}$, $\hat{\theta}]=G(J)$. These predictions are supervised by the ground truth shape and pose parameters. To further ensure that the predicted $\hat{\beta}$ and $\hat{\theta}$ are accurate, we also introduce a joints-cycle-consistency learning objective which ensures the 3D keypoints inferred from the predicted $\hat{\beta}$ and $\hat{\theta}$ are close to the input 3D keypoints. Mathematically, we seek to enforce a forward cycle consistency as: $J \rightarrow G(J) \rightarrow X(G(J)) = \hat{J} \approx J$. The overall loss function to train our USCG is:
\begin{equation}\label{eq:update}
\begin{aligned}
    L = & \lVert [\beta,\theta] -[\hat{\beta},\hat{\theta}] \rVert_2^2
        + \lVert J - X(G(J)) \rVert_2^2.
\end{aligned}
\end{equation}

Our USCG module is trained in this fashion with the MoSh data described above. Once trained, the USCG will provide our overall training pipeline of Figure~\ref{fig:pipeline} with auxiliary SMPL training constraints by generating $[\tilde{\beta},\tilde{\theta}]$ from the 3D keypoints information available as part of RGB-D datasets. The predicted $[\tilde{\beta},\tilde{\theta}]$ will then be used to provide supervisory signals to constrain the shape and pose estimated by our model. Note that before using these estimated constraints, we perform a quality check to filter out any obviously incorrect $[\tilde{\beta},\tilde{\theta}]$ predictions. To realize this, we set a threshold on the reconstruction error, which is the mean per joint position error (MPJPE) after rigid alignment via Procrustes Analysis \cite{gower1975generalized}, between $J$ and $\hat{J}$ after the forward cycle.

Once the SMPL constraints are generated by our trained USCG, we use the standard Euclidean distance metric to optimize the SMPL learning objective when training our proposed RGB-D model:
\begin{equation}\label{eq:lossUscg}
L_{smpl} =\begin{matrix} \sum_{i} \lVert [\tilde{\beta},\tilde{\theta}] -[\hat{\beta},\hat{\theta}] \rVert_2^2 \end{matrix},\\
\end{equation}

\subsection{Depth Ranking Consistency}
\label{sec:rankLoss}
Given $\theta$ and $\beta$ predictions, we can use the SMPL forwarding functionality to infer the 3D joints positions $(x_{J},y_{J},z_{J})$. Along with our RGB-D pipeline and the constraint generator, we also propose a new learning objective we call depth ranking consistency to address the depth ambiguity problem. The intuition of the depth ranking consistency is to ensure the relative ordering of the predicted joints locations is consistent with the input joints. Given an input 2D keypoint $(x,y)$, we obtain its corresponding depth $z_{d}$ from the aligned depth map. While the raw depth value $z_{d}$ is not directly comparable to the $z_{J}$ obtained from the SMPL forwarding function (as the coordinate system definitions are different), the relative depth orderings (\ie, closer-farther) between each pair of joints $z_{J}$ and $z_{d}$ will have to be consistent. To this end, our depth ranking consistency learning objective explicitly enforces our network to predict $\theta$ and $\beta$ that satisfies this relative depth ordering property. Specifically, for a pair of joints $(p,q)$, we define its depth ranking relation $r_{p,q}$ as:
\begin{equation}
r_{p,q}=\begin{cases} 
  1, & \mbox{if } z_{d}^p < z_{d}^q\\
 -1, & \mbox{if } z_{d}^p > z_{d}^q\\
  0, & \mbox{if } z_{d}^p = z_{d}^q.
\end{cases}
\end{equation}
The depth ranking consistency objective penalizes the case when a pair of the inferred 3D joints, from the predicted $\theta$ and $\beta$, has relative depth relationship that is opposite to the relationship derived from the input depth. Our objective function can be expressed as:
\begin{equation}\label{eq:lossDrc}
L_{drc} =\sum_{(p,q)\in P} L_{p,q}
\end{equation}
where $P$ represents the set containing the non-repetitive pairs of joints and $L_{p,q}=\text{log}(1 + \text{exp}(r_{p,q}(z_{J}^p - z_{J}^q))$. 

\subsection{Overall Learning Objective}
Our proposed overall objective function is:
\begin{equation}\label{eq:loss}
\begin{aligned}
    L & = \lambda_{2d}L_{2d} + \lambda_{smpl}L_{smpl} + \lambda_{drc}L_{drc} + \lambda_{adv}L_{adv},
\end{aligned}
\end{equation}
where $L_{2d}$ is an L1 norm loss on the 14 2D joints as defined in prior work \cite{Bogo:ECCV:2016, kanazawaHMR18,lassner2017unite}, $L_{adv}$ is the same discriminator loss of Kanazawa \etal \cite{kanazawaHMR18}, $L_{smpl}$ and $L_{drc}$ are in Equations~\ref{eq:lossUscg} and~\ref{eq:lossDrc}, and the $\lambda$s are the corresponding loss weights.

\section{Implementation Details}
\noindent
\textbf{Datasets} \, We use PKU-MMD \cite{pku-mmd} and CAD-60 \cite{cad60} as our RGB-D datasets. They provide aligned RGB-D images and 3D joints annotations. PKU-MMD covers a wide range of complex human activities while CAD-60 has a relatively small number of activities (12 in total). Both these datasets do not have SMPL annotations. 
To expand the variety of the training data, we also use RGB datasets that provide only 2D joints annotations: MS COCO \cite{coco} and LSP \cite{lsp}. 
\noindent
\textbf{Architecture} \,We use two separate ResNet-50 networks, without the top fully connected (FC) layers, as RGB and depth encoders, giving $\phi_{RGB} \in \mathbb{R}^{2048}$ and $\phi_{D} \in \mathbb{R}^{2048}$. Our fusion module has two FC layers (with 4086 and 2048 units), each with ReLU activation, with $p_{\text{miss}}=0.3$. The USCG is implemented with a stack of 10 FC layers (with ReLU activations) and 100mm as the quality-check threshold $e$. When training with RGB-only and RGB-D datasets, we ensure each batch contains data samples equally distributed among the two. The batch size is set to 40, learning rate to $10^{-5}$, and our code is written in PyTorch \cite{pytorch}.

\section{Experiments and Results}

\begin{figure*}[t]
  \centering
  \subfigure[$\text{HMR}_{\text{RGBD}}$]{\label{fig:naive}
  \includegraphics[width=5.5cm]{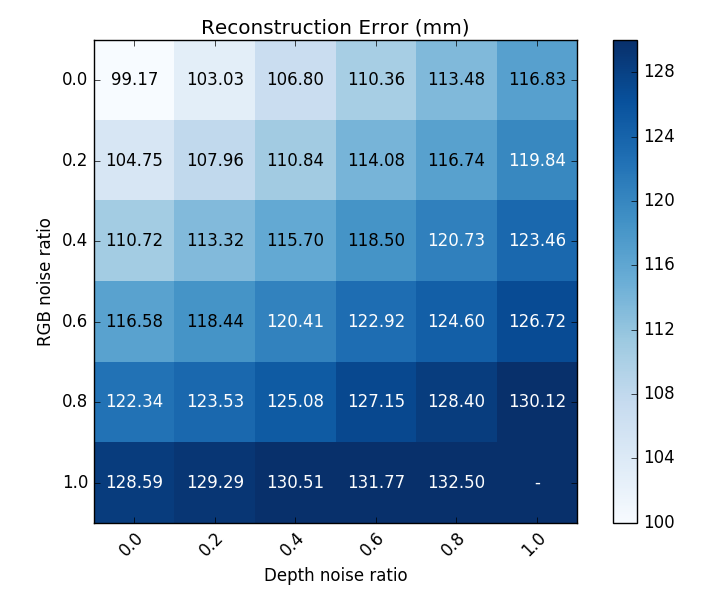}}
  \subfigure[Ours]{\label{fig:ours}
  \includegraphics[width=5.5cm]{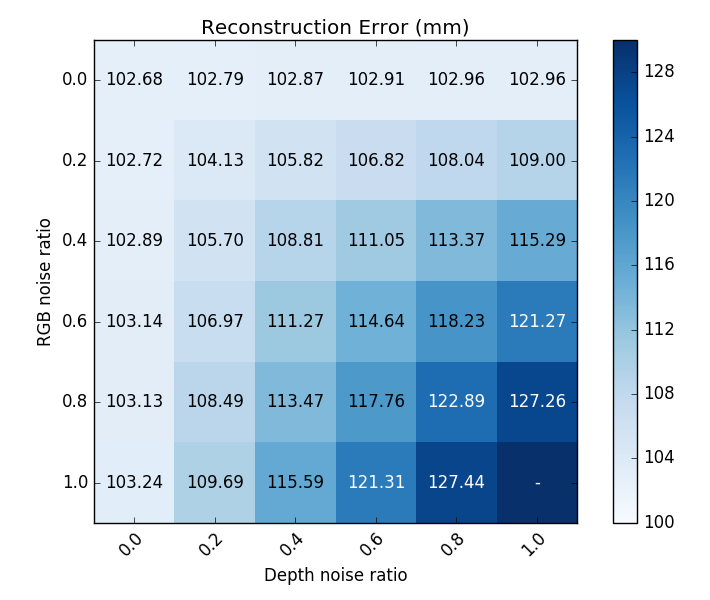}}
  \subfigure[Performance gain of ours]{\label{fig:diff}
  \includegraphics[width=5.5cm]{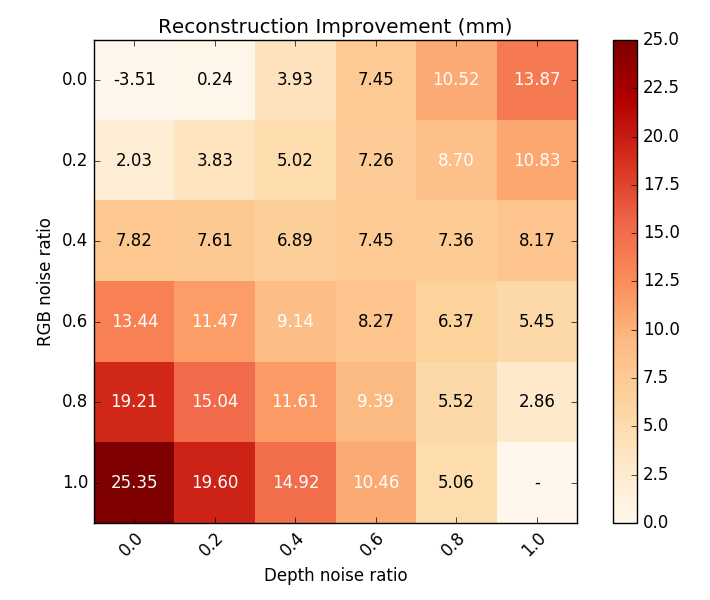}}
  \caption{Comparison under different noise ratios. (a) $\text{HMR}_{\text{RGBD}}$. (b) Ours. (c) Our improvement $(a)-(b)$.}\label{fig:cm}
\end{figure*}

\begin{table*}[!t]
\centering
\begin{tabular}{c|>{\centering\arraybackslash}p{2cm} >{\centering\arraybackslash}p{2cm}|>{\centering\arraybackslash}p{2cm} >{\centering\arraybackslash}p{2cm}|>{\centering\arraybackslash}p{2cm} >{\centering\arraybackslash}p{2cm}}
\hline
\multirow{2}{*}{Input} & \multirow{2}{*}{\tabincell{c}{$\text{HMR}_{\text{RGB}}$}} & \multirow{2}{*}{\tabincell{c}{$\text{HMR}_{\text{D}}$}} &  $\text{HMR}_{\text{RGBD}}$    & $\text{HMR}_{\text{RGBD}}$ & \textbf{Fusion}     & \textbf{Fusion}  \\
                       &                                             &                                           &  no DRC & with DRC   & no DRC   & with DRC     \\ 
\hline
RGB                    & 133.70                                      & -                                         & -          & -           & 139.02     & \textbf{111.88}       \\
D                      & -                                           & 173.69                                    & -          & -           & 141.93     & \textbf{128.06}       \\
RGB-D                  & -                                           & -                                         & 136.18     & 122.65      & 136.61     & \textbf{107.86}       \\
\hline
\end{tabular}
\caption{Reconstruction errors (mm) on PKU-MMD. Lower is better.}\label{tab:PKU}
\end{table*}

\begin{table*}[t]
\centering
\begin{tabular}{c|>{\centering\arraybackslash}p{2cm} >{\centering\arraybackslash}p{2cm}|>{\centering\arraybackslash}p{2cm} >{\centering\arraybackslash}p{2cm}|>{\centering\arraybackslash}p{2cm} >{\centering\arraybackslash}p{2cm}}
\hline
\multirow{2}{*}{Input} & \multirow{2}{*}{\tabincell{c}{$\text{HMR}_{\text{RGB}}$}} & \multirow{2}{*}{\tabincell{c}{$\text{HMR}_{\text{D}}$}} &  $\text{HMR}_{\text{RGBD}}$    & $\text{HMR}_{\text{RGBD}}$ & \textbf{Fusion}     & \textbf{Fusion}  \\
                       &                                             &                                           &  no DRC & with DRC   & no DRC   & with DRC     \\ 
\hline
RGB                    & 119.22                                      & -                                         & -          & -                 & 106.99     & \textbf{102.96}       \\
D                      & -                                           & 108.98                                    & -          & -                 & 109.47     & \textbf{103.24}       \\
RGB-D                  & -                                           & -                                         & 101.43     & \textbf{99.17}    & 106.96     & 102.68       \\
\hline
\end{tabular}
\caption{Reconstruction errors (mm) on CAD-60. Lower is better.}\label{tab:CAD}
\end{table*}

We evaluate the efficacy of our proposed dynamic data fusion module, the universal SMPL constraint generator module, and the depth ranking consistency loss thoroughly by a variety of experiments on the RGB-D datasets considered in this work. In all our experiments, as noted in Section~\ref{sec:dynamicFusion}, we use the HMR algorithm of Kanazawa \etal \cite{kanazawaHMR18} as our baseline, and denote the HMR model trained with RGB-only and depth-only input as $\text{HMR}_{\text{RGB}}$ and $\text{HMR}_{\text{D}}$ respectively. $\text{HMR}_{\text{RGBD}}$ denotes the model that uses a direct concatenation as the fusion method when trained on RGB-D input. In other words, $\text{HMR}_{\text{RGBD}}$ represents the scenario where we have both RGB and depth data available during both training and inference. We test all baselines with the same kind of input as used in training. For example, $\text{HMR}_{\text{RGB}}$ and $\text{HMR}_{\text{D}}$ are tested on RGB and depth images only, respectively, whereas $\text{HMR}_{\text{RGBD}}$ is tested on the aligned RGB-D data. Our proposed model, represented by $\text{Fusion}$, is tested under 3 scenarios (to demonstrate our fusion): RGB-only input, depth-only input, and RGB-D input, denoted RGB, D, and RGB-D respectively in the tables that follow. We use the reconstruction error (MPJPE after rigid alignment as noted previously) as the evaluation metric.

\begin{table*}[t]
\centering

\begin{tabular}{c|>{\centering\arraybackslash}p{1.8cm} >{\centering\arraybackslash}p{1.8cm}|>{\centering\arraybackslash}p{1.8cm} >{\centering\arraybackslash}p{1.8cm} cc}
\hline
\multirow{3}{*}{Input} & RGB-D              & RGB-D              &  RGB-D     & RGB-D       & RGB-D            & RGB-D             \\
                       & 2d+adv+3D          & 2d+adv+3D          &  2d+adv    & 2d+adv+smpl & 2d+adv+rank      & 2d+adv+smpl+rank  \\
                       & PKU                & CAD                &  PKU+CAD   & PKU+CAD     & PKU+CAD          & PKU+CAD           \\ 
\hline
RGB (PKU)              & \textbf{96.70}     & 131.63             & 132.19     & 126.64      & 122.25           & \textbf{120.98}   \\
D (PKU)                & \textbf{100.64}    & 142.10             & 132.74     & 131.43      & \textbf{121.08}  & 123.29            \\
RGB-D (PKU)            & \textbf{96.62}     & 131.21             & 130.70     & 126.74      & 120.90           & \textbf{120.59}   \\
\hline
RGB (CAD)              & 133.77             & \textbf{79.45}     & 126.08     & 111.37      & 105.51           & \textbf{102.76}   \\
D (CAD)                & 134.34             & \textbf{80.64}     & 126.28     & 113.18      & 105.70           & \textbf{103.45}   \\
RGB-D (CAD)            & 132.31             & \textbf{79.33}     & 124.67     & 111.46      & 104.87           & \textbf{102.45}   \\
\hline
\end{tabular}
\caption{Reconstruction errors (mm) on PKU-MMD and CAD-60. All the models are trained with our dynamic fusion module.}\label{tab:PKUCAD}
\end{table*}

\subsection{Impact of Dynamic Data Fusion}
To evaluate the robustness of our fusion module, we progressively increase the level of noise (specifically, we increase the probability $p_{\text{miss}}$ from 0.0 to 1.0) in both the RGB and depth streams during inference on the CAD-60 dataset, and compare the performance of our method with the baseline $\text{HMR}_{\text{RGBD}}$. Figure \ref{fig:cm}(a) shows a heatmap representation of the reconstruction error of $\text{HMR}_{\text{RGBD}}$, whereas Figure \ref{fig:cm}(b) shows our results. As can be observed from the figures, with an increase in the level of noise, the performance of both models goes down as expected. However, as indicated by Figure \ref{fig:cm}(c), our proposed model suffers less degradation when compared to $\text{HMR}_{\text{RGBD}}$, suggesting better robustness with partially missing aligned RGB-D data. 

Tables~\ref{tab:PKU} and~\ref{tab:CAD} show the reconstruction error on PKU-MMD and CAD-60 datasets respectively. We make several observations. We note that the performance of our proposed fusion method (trained with probabilistically missing RGB-D data) is not expected to be better than the baseline $\text{HMR}_{\text{RGBD}}$ (which is trained RGB and D both available), but we see from column 4 and 6 that with RGB-D input, our error is 136.61 (quite close to baseline 136.18). However, with our proposed RGB-D fusion, we can now achieve better performance (error of 141.93) compared to the $\text{HMR}_{\text{D}}$ baseline that is trained only on depth data. Since our fusion results in column 6 are with only $L_{2d}$ and $L_{adv}$ losses (\ie, no $L_{drc}$ or $L_{smpl}$), we do not expect much improvement when tested with RGB-only input. Despite this, our fusion achieves reasonably competitive performance (139.02 error) when compared to $\text{HMR}_{\text{R}}$, the RGB-only baseline, which achieves 133.7. However, to realize the full value of our fusion pipeline, we test our fusion pipeline with $L_{drc}$, since this will enable us to provide more explicit 3D supervision. Results with this configuration are shown in column 7, where we see consistent performance improvement across all the input types. It is also worth pointing out that our fusion with DRC gives better performance than the baseline $\text{HMR}_{\text{RGBD}}$ with DRC. 

\subsection{Impact of Depth Rank Consistency}
Columns 4, 5, 6, and 7 in Tables~\ref{tab:PKU} and~\ref{tab:CAD} show results of both baseline HMR and our pipeline with and without the proposed DRC learning objective. As can be seen from the results, adding $L_{drc}$ to the training objective results in a decrease in the reconstruction error. Notably, the error of our model with the RGB-D input on the PKU-MMD data drops by over 30mm. We further note that the error reduction is not limited to only our method, but even the baseline HMR benefits from the proposed $L_{drc}$ loss. Additional experiments to further demonstrate the impact of DRC are discussed in Section~\ref{sec:resUSCG} below.

\subsection{Impact of USCG}
\label{sec:resUSCG}
As noted previously, each dataset typically has 3D joints annotations in their specific coordinate system, and training a model, with direct 3D joints supervision, on one dataset does not necessarily result in good generalization on another dataset. To understand this aspect better, we train two $\text{HMR}_{\text{RGBD}}$ models with the 3D joints loss and our fusion module (denoted 2d+adv+3D in Table~\ref{tab:PKUCAD}) on the PKU-MMD and CAD-60 datasets separately. We summarize testing results on each dataset in Table~\ref{tab:PKUCAD}. As can be seen from the results, both models perform worse when tested on a dataset they were not trained on. For instance, the PKU-MMD model trained with RGB-D data on the PKU-MMD data gives a reconstruction error of 96.62 on the PKU-MMD data, while giving 131.21 on the CAD data. A similar observation holds for the model trained on the CAD-60 data as well. To address this issue, as motivated in Section~\ref{sec:uscg}, we train a model combining both datasets. However, naive combination (with the 3D joints loss) of the two datasets is not meaningful since the 3D joints information of each dataset is in a different coordinate system. To address this issue and provide our model with meaningful 3D supervision, we generate auxiliary SMPL constraints using our proposed USCG module of Section~\ref{sec:uscg}. We take the 3D joints provided by datasets, use our USCG to produce the corresponding SMPL parameters, and retain only those parameters that pass our quality check described in Section~\ref{sec:uscg}. Once these constraints are obtained, we train a model on both PKU-MMD and CAD-60 datasets with the $L_{smpl}$ loss as well. To understand the impact of this auxiliary $L_{smpl}$ loss, we refer to column 4 and column 5 in Table~\ref{tab:PKUCAD}. Column 4 shows the performance of the baseline $\text{HMR}_{\text{RGBD}}$ model with our fusion module, whereas column 5 shows its performance when also trained with our auxiliary $L_{smpl}$ loss. We clearly note a consistent reduction in the reconstruction error across all the three input types (RGB-only, Depth-only, and RGB-D) and both datasets (PKU-MMD and CAD-60). To more explicitly use 3D joints during training as well, we incorporate our proposed DRC. Note that while 3D joints for the two datasets may be in different coordinate systems, our proposed DRC objective only looks at the relative ordering configuration of the 3D joints, helping alleviate this problem. We summarize our results in columns 6 and 7 of Table~\ref{tab:PKUCAD}. In column 6, we add our proposed DRC to the baseline $\text{HMR}_{\text{RGBD}}$, resulting in substantial reductions in reconstruction error across all input types and both datasets (\eg, 130.70 with baseline $\text{HMR}_{\text{RGBD}}$ on RGBD input and 120.90 when proposed DRC is added to baseline $\text{HMR}_{\text{RGBD}}$). These error reductions suggest that our proposed DRC is able to account for the coordinate system differences and help improve the baseline performance. Finally, we integrate both $L_{smpl}$ and $L_{drc}$ into the baseline $\text{HMR}_{\text{RGBD}}$ (results in column 7 of Table~\ref{tab:PKUCAD}), which gives the better performance than using either $L_{smpl}$ or $L_{drc}$ by itself in most cases. We would also like to note that while results may not be directly comparable to the case when the model is trained and tested on the same dataset (whose performance is naturally expected to be the best), we observe our model trained on both datasets with USCG and DRC and tested on each individual dataset performs better than training the baseline model on one dataset and testing on another dataset. These results suggest our proposed USCG and DRC has the potential to enable unified model training across multiple RGB-D datasets despite variations in thier coordinate system definitions.

\subsection{Training with RGB Datasets}
While RGB datasets cannot provide direct 3D supervisory signals to address the depth ambiguity problem, their 2D keypoints annotations can certainly help improve model generalizability. With our proposed dynamic data fusion module, we can now integrate RGB-D and RGB datasets that provide different kinds of annotations to train a unified model. To this end, we follow the same workflow described in our fusion module in Section~\ref{sec:dynamicFusion}, but now also have RGB-only datasets in addition to RGB-D datasets. For training, with data from the RGB-only datasets, we simply pair the RGB images with zero-value blank images to make up the missing depth channel, following our discussion in Section~\ref{sec:dynamicFusion}. We now train our proposed model (with $L_{smpl}$ and $L_{drc}$) in two ways: (a) with a combination of RGB-D and RGB-only datasets, and (b) with only RGB-D datasets. We summarize results in Table~\ref{tab:coco}, where we see training with the additional RGB-only data gives consistent reduction in reconstruction errors across all input types and both the testing datasets. These results further demonstrate the impact of our fusion module in training generalizable models with different kinds of datasets each of which provides different kinds of annotations. 
\begin{table}[t]
\centering
\small
\begin{tabular}{c|>{\centering\arraybackslash}p{2.5cm} >{\centering\arraybackslash}p{2.5cm}}
\hline
\multirow{3}{*}{Input} & RGB-D                & RGB-D            \\
                       & 2d+adv+smpl+rank     & 2d+adv+smpl+rank \\
                       & w/o RGB datasets     & w/ RGB datasets   \\ 
\hline
RGB (PKU)              & 120.98               & \textbf{111.56}  \\
D (PKU)                & 123.29               & \textbf{112.85}  \\
RGB-D (PKU)            & 120.59               & \textbf{110.47}  \\
\hline
RGB (CAD)              & 102.76               & \textbf{93.92}   \\
D (CAD)                & 103.45               & \textbf{91.13}   \\
RGB-D (CAD)            & 102.45               & \textbf{91.04}   \\
\hline
\end{tabular}
\caption{Results w/ and w/o RGB datasets.}\label{tab:coco}
\end{table}

\section{Summary}
We proposed a new technique to address the human mesh estimation problem with RGB-D data. We presented a dynamic data fusion module that gives improved estimation robustness with scenarios involving potentially missing aligned RGB-D data. To realize this, we also presented an associated probabilistic training policy to simulate this aspect of missing data, which also helps us train models with a wide variety of datasets. To deal with the wide variety of available or unavailable annotations, we presented a new SMPL constraint generator that helps provide auxiliary SMPL supervisory signals for RGB-D datasets that have no explicit SMPL annotations. Finally, to ensure depth consistent parameter estimation, we proposed a new depth ranking consistency loss to guide the model to respect the relative depth ordering of predicted keypoints. 

{\small
\bibliographystyle{ieee_fullname}
\bibliography{ref}
}

\end{document}